\documentclass{bmvc2k}

\title{Deep Least Squares Alignment for Unsupervised Domain Adaptation}

\addauthor{Youshan Zhang}{https://sites.google.com/view/youshanzhang}{1}
\addauthor{Brian D.\ Davison}{http://www.cse.lehigh.edu/~brian/}{1}

\addinstitution{
Computer Science and Engineering\\
Lehigh University\\
Bethlehem, PA, USA
}

\runninghead{Y. Zhang and B. D.\ Davison}{Deep Least Squares Alignment for UDA}

\usepackage{hyperref}       
\usepackage{url}            
\usepackage{booktabs}       
\usepackage{amsfonts}       
\usepackage{nicefrac}       
\usepackage{microtype}      
\usepackage{xcolor}         
\usepackage{mathrsfs}  
\usepackage{bm}
\usepackage{amssymb,amsmath}
\usepackage{multicol,multicap,graphicx}
\usepackage{graphicx}
\usepackage{caption}
\usepackage{wrapfig}
\usepackage{picinpar}
\usepackage{cutwin}
\usepackage{stmaryrd}
\usepackage{graphicx,multicol}
\usepackage{algorithm}  
\usepackage{algorithmic}
\usepackage{amsfonts}
\usepackage{amsfonts,amssymb} 
\usepackage{verbatim}
\usepackage{url}
\usepackage{enumitem}

\DeclareMathOperator*{\argmin}{arg\,min}

\newcommand{\overbar}[1]{\mkern 3.5mu\overline{\mkern-3.5mu#1\mkern-3.5mu}\mkern 3.5mu}

\begin{document}

\maketitle

\begin{abstract}
Unsupervised domain adaptation leverages rich information from a labeled source domain to model an unlabeled target domain. Existing methods attempt to align the cross-domain distributions. However, the statistical representations of the alignment of the two domains are not well addressed.  In this paper, we propose deep least squares alignment (DLSA) to estimate the distribution of the two domains in a latent space by parameterizing a linear model. We further develop marginal and conditional adaptation loss to reduce the domain discrepancy by minimizing the angle between fitting lines and intercept differences and further learning domain invariant features. Extensive experiments demonstrate that the proposed DLSA model is effective in aligning domain distributions and outperforms state-of-the-art methods.
\end{abstract}

\section{Introduction}
Large amounts of labeled data is a prerequisite to training accurate predictors in most machine learning techniques. However, manually labeling and training a model from scratch is tedious and expensive. Fortunately, unsupervised domain adaptation (UDA) aims to deal with the shortage of labels by leveraging a richly labeled source domain to a similar but different unlabeled target domain. This task is usually challenged by the dataset bias or domain shift issue because source and target domains have different characteristics. UDA can mitigate this 
by establishing the association between 
domains and learning domain invariant features.


Recent advances in UDA witness its success in deep neural networks. It can learn abstract representations with nonlinear transformations and suppress the negative effects caused by the domain shift. In earlier work, deep learning based methods rely on minimizing the discrepancy between the source and target distributions by proposing different loss functions, such as Maximum Mean Discrepancy (MMD)~\cite{tzeng2014deep}, CORrelation ALignment (CORAL)~\cite{sun2016deep}, Kullback-Leibler divergence (KL)~\cite{meng2018adversarial}. Inspired by generative adversarial network (GAN) ~\cite{goodfellow2014generative}, adversarial domain adaptation methods aim to identify domain invariant features by playing a min-max game between domain discriminator and feature extractor~\cite{ghifary2014domain,tzeng2017adversarial,zhang2020adversarial1,zhang2021efficient,zhang2020adversarial3}. However, these methods either cannot fully align the marginal and conditional distributions between two domains or request additional components such as a domain discriminator~\cite{tzeng2017adversarial} or gradient reversal layer~\cite{ghifary2014domain}. 
 
\begin{wrapfigure}[19]{r}{0.5\textwidth}
  \begin{center}
  \captionsetup{font=small}
  \vspace{-.8cm}
    \includegraphics[width=0.48\textwidth]{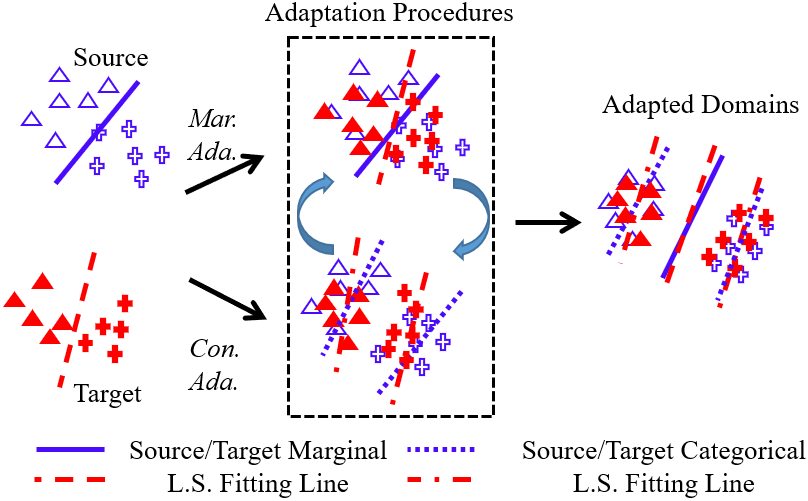}
  \end{center}
  \caption{Schematic diagram of the proposed least squares alignment. We first estimate the least squares fitting lines of two domains and then align the marginal and conditional distributions. By rotating and translating the fitting lines, the distributions between the two domains can be aligned, and domain discrepancy is minimized (Mar.: marginal, Con.: conditional, and Ada.: adaptation).} \label{fig:shem}
\end{wrapfigure}

Although many methods achieve remarkable results in domain adaptation, they still suffer from two challenges: 1) the distributions of two domains cannot be intuitively represented, and alignment processes are hidden; and, 2) the label information and latent structure of the target domain are not fully considered, and how to better align marginal and conditional distributions are not well addressed. To alleviate these challenges, we propose a deep least squares adaptation (DLSA) model with a toy example shown in Fig.~\ref{fig:shem}. 

We offer two contributions:

\begin{enumerate}
\vspace{-0.2cm}
    \item We propose a simple and novel UDA approach, DLSA, to explicitly model the distribution of domains in the latent space with estimated fitting lines, which are parameterized by estimated slope and intercept. We further theoretically and statistically show the effectiveness of DLSA in estimating domain distributions.

    \item We design and effectively integrate marginal and conditional adaptation losses to impose distribution alignment. By minimizing angle and intercept differences between source and target fitting lines, we enforce feature discriminability, which leads to inter-class dispersion and intra-class compactness.
\end{enumerate}
\vspace{-0.2cm}
Experimental results on three benchmark datasets show that DLSA achieves higher classification performance than state-of-the-art methods. We also statistically show the estimated least squares parameters to model the distributions of source and target domains.

\section{Related work}
There are different deep techniques for UDA. To learn domain invariant representations, early methods attempt to propose distance-based loss functions to align data distributions between different domains. Maximum Mean Discrepancy (MMD)~\cite{tzeng2014deep} is one of the most popular distance functions to minimize between two distributions. Deep Adaptation Network (DAN) considered the sum of MMD from several layers with multiple kernels
of MMD functions~\cite{long2015learning}. The CORAL loss is another distance function, based on covariance matrices of the latent features from two domains~\cite{sun2016deep}. Recently, Li et al.~\cite{li2020enhanced} proposed an Enhanced Transport Distance (ETD) to measure domain discrepancy by establishing the transport distance of attention perception as the predictive feedback of iterative learning classifiers. 

With the advent of GAN~\cite{goodfellow2014generative},  adversarial learning models have been found to be an impactful mechanism for identifying invariant representations in domain adaptation and minimizing the domain discrepancy.  The Domain Adversarial Neural Network (DANN) considered a minimax loss to integrate a gradient reversal layer to promote the discrimination of source and target domains~\cite{ganin2016domain}.  The Adversarial Discriminative Domain Adaptation (ADDA) method utilized an inverted labeled GAN loss to split the source and target domains, and features can be learned separately~\cite{tzeng2017adversarial}. Xu et al.~(\cite{xu2019adversarial}) mapped the two domains to a common potential distribution and effectively transfers domain knowledge. There are also many methods that utilized pseudo-labels to consider label information in the target domain~\cite{chen2019progressive,kang2019contrastive,zhang2021adversarial2}. They have not, however, intuitively studied the distribution adaptation as thoroughly as we do in Fig.~3 of supplemental material.

Least squares estimation is also used in domain adaptation, but most are proposed to solve regression problems. Huang et al.~\cite{huang2020domain} proposed a domain adaptive partial least squares regression model, which utilized the Hilbert-Schmidt independence criterion to estimate the independence of the extracted latent variables and domain labels. The partial least squares method was used to align the source and target data in the latent space via estimating a projection matrix. Similarly, Nikzad-Langerodi et al.~\cite{nikzad2020domain} considered UDA for regression problems under Beer–Lambert’s law. They employed a non-linear iterative partial least squares algorithm to minimize the covariance matrices difference of the latent sample between two domains. Yuan et al.~\cite{yuan2020adversarial} proposed to use least squares distance to align marginal distribution between two domains for classification problem. However, the so-called least-squares distance is proposed by Mao et al.~\cite{mao2017least}, aiming to push generated samples toward the decision boundary and reduce the gradient vanishing problem during adversarial learning. Notably, we focus on UDA for visual recognition and impose marginal and conditional distribution adaptation losses based on slope and intercept differences from the least squares estimation.

\section{Methodology}
\subsection{Problem and motivation}
For unsupervised domain adaptation, given a source domain $\mathcal{D_S} = \{\mathcal{X}_\mathcal{S}^i, \mathcal{Y}_\mathcal{S}^i \}_{i=1}^{\mathcal{N}_\mathcal{S}}$ of $\mathcal{N}_\mathcal{S}$ labeled samples in $C$ categories and a target domain $\mathcal{D_T} = \{\mathcal{X}_\mathcal{T}^j\}_{j=1}^{\mathcal{N}_\mathcal{T}}$ of $\mathcal{N}_\mathcal{T}$ samples without any labels (\textit{i.e.,} $\mathcal{Y}_\mathcal{T}$ is unknown). Our ultimate goal is to learn a classifier $\mathcal{F}$ under a feature extractor $G$, which reduces domain discrepancy and improves the generalization ability of the classifier to the target domain. 

To achieve it, existing methods usually attempt to align either marginal or conditional distributions of the two domains. Moreover, the statistical estimation of distributions is not well addressed. In contrast, we propose to align both distributions to further reduce domain discrepancies. In particular, we employ least squares to estimate the latent space distribution, which is parameterized by a slope and an intercept. We then design marginal and conditional adaptation losses to enforce the distribution alignment both in inter-class  and intra-class. In turn, we can push the decision boundary of classifier $\mathcal{F}$  toward the target domain.

\begin{figure*}[t]
\centering
\captionsetup{font=small}
\includegraphics[scale=0.24]{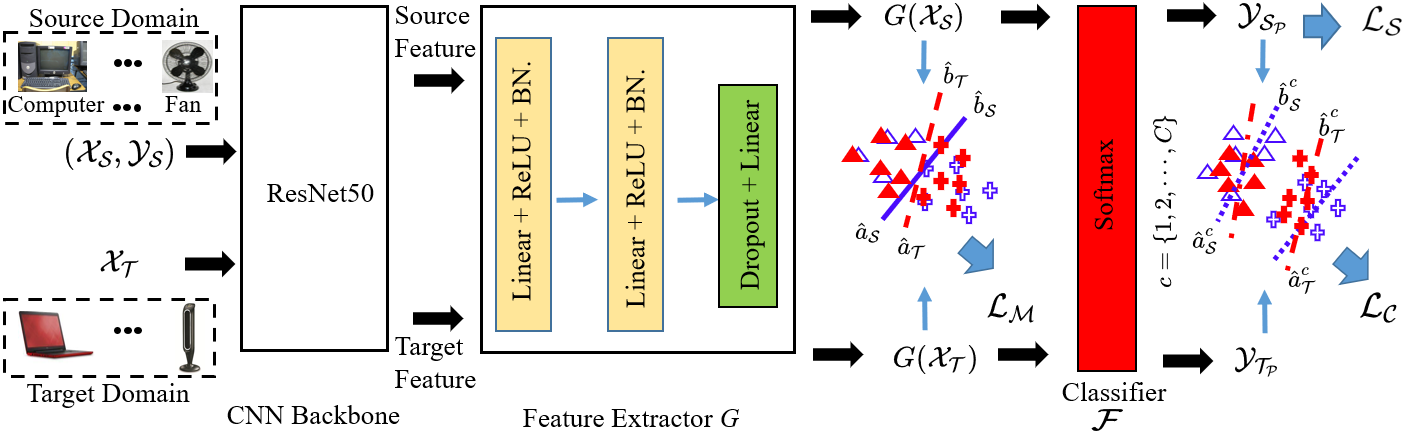}
\vspace{+0.15cm}
\caption{The architecture of our proposed DLSA model. We first employ a ResNet50-based feature extractor $G$ to acquire the feature representations of two domains. DLSA consists of three loss functions (classification loss $\mathcal{L_S}$, marginal adaptation loss  $\mathcal{L_M}$, and conditional adaptation loss $\mathcal{L_C}$). The marginal distribution is aligned via reducing the global angle and intercept differences between source and target fitting lines. The conditional distribution is aligned by categorical angle and intercept differences of two domains (BN.: BatchNormalization). }
\label{fig:model}
\end{figure*}

\vspace{-0.3cm}
\subsection{Source Classifier}
For the labeled source domain, we minimize the following classification loss:
\begin{equation}\label{eq:lc}
    \mathcal{L_S} =J(\mathcal{F}(G(\mathcal{X}_{\mathcal{S}})), \mathcal{Y_S}),
\end{equation}
where $\mathcal{F}$ is the classifier, $J(\cdot , \cdot)$ is the cross-entropy loss and $\mathcal{Y_{S_P}} = \mathcal{F}(G(\mathcal{X}_{\mathcal{S}}))$ is the predicted label in Fig.~\ref{fig:model}.


\subsection{Least squares estimation}\label{sec:ls}
The feature extractor $G$ maps samples from a given neural network layer into a $d$-dimensional latent space ($d>1$). The latent source and target samples can be denoted as: $G(\mathcal{X_S}) \in \mathbb{R}^{\mathcal{N_S}\times d}$, and $G(\mathcal{X_T}) \in \mathbb{R}^{\mathcal{N_T}\times d} $, respectively. Next, we aim to model domain distribution via estimating the fitting line of two domains in the latent space, which has the minimum error to all latent samples.  The latent samples can be encoded via $G(\mathcal{X_{Z}}) = [G(\mathcal{X_{Z}})^1, G(\mathcal{X_{Z}})^{d-1}]$, where $\mathcal{Z}$ can be either source domain $\mathcal{S}$ or target domain $\mathcal{T}$. Let $L_V^\mathcal{Z}= G(\mathcal{X_{Z}})^1 \in \mathbb{R}^{\mathcal{N_Z} \times 1}$ and $L_W^\mathcal{Z} = G(\mathcal{X_{Z}})^{d-1} \in \mathbb{R}^{\mathcal{N_Z} \times (d-1)}$, $L_V^z$ and $L_W^z$ are each one of the elements, respectively, \textit{i.e.,} $L_V^z \in \mathbb{R}$, which is a scalar and $L_W^z \in \mathbb{R}^{d-1} (z=1, 2, \cdots, \mathcal{N_Z})$. To model the distribution of latent space, we naively assume that there is a linear relationship between $L_V^\mathcal{Z}$ and $L_W^\mathcal{Z}$. The motivation is that we aim to use the slope and intercept to represent the distribution of the latent feature $G(\mathcal{X_Z})$. Then, we can minimize the difference of slope and intercept of the two domains. Moreover, such a linear fitting line can always be estimated (except when there is only one sample for estimation). Here, we arbitrarily choose the first dimension as the independent variable, and the remaining $d-1$ dimensions as dependent variables; other dimensions are explored in supplemental material. 

Before formulating distribution alignment, we first review multiple linear regression in $\mathbb{R}^{d-1}$. Given the target variable (dependent variable) $L_W^\mathcal{Z}$ taking values in $\mathbb{R}^{d-1}$, and independent variable $L_V^\mathcal{Z}$ taking values in $\mathbb{R}$, the multiple linear regression model is given by:
\begin{equation}\label{lg}
    L_W^\mathcal{Z} = a_\mathcal{Z} L_V^\mathcal{Z} +b_\mathcal{Z} +\epsilon,
\end{equation}
where $a_\mathcal{Z} \in \mathbb{R}^{d-1}$ contains the unobservable \textit{slope} parameter, $b_\mathcal{Z} \in \mathbb{R}^{d-1}$ holds unobservable \textit{intercept} parameter, and $\epsilon$ is unobservable random noise, which is drawn  \textit{i.i.d}. 

Consider a data set of input $L_V^\mathcal{Z}=\{L_V^z \}_{z=1}^{\mathcal{N_Z}}$ with corresponding target value $L_W^\mathcal{Z}=\{L_W^z \}_{z=1}^{\mathcal{N_Z}}$, the least squares estimate of the slope $\hat{a}_{\mathcal{Z}}$ and intercept $\hat{b}_{\mathcal{Z}}$  via solving the minimization problem in Eq.~\ref{eq:ls}.
\begin{equation}\label{eq:ls}
    \hat{a}_{\mathcal{Z}}, \hat{b}_{\mathcal{Z}} = \arg \min \sum_{z=1}^{\mathcal{N_Z}}||L_W^z - a_{\mathcal{Z}} L_V^z - b_{\mathcal{Z}}||^2
\end{equation}
Let $E = \sum_{z=1}^{\mathcal{N_Z}}||L_W^z - a_{\mathcal{Z}} L_V^z - b_{\mathcal{Z}}||^2$, to estimate coefficients, we take the gradient of $E$ with respect to each parameter, and setting the result equal to zero.
\begin{equation}
\begin{aligned}\label{eq:par}
& \frac{\partial E}{\partial a_{\mathcal{Z}}} = -2 \sum_{z=1}^{\mathcal{N_Z}} L_V^z(L_W^z - a_{\mathcal{Z}} L_V^z - b_{\mathcal{Z}}) = 0, \quad \frac{\partial E}{\partial b_{\mathcal{Z}}} = -2 \sum_{z=1}^{\mathcal{N_Z}} (L_W^z - a_{\mathcal{Z}} L_V^z - b_{\mathcal{Z}}) = 0
\end{aligned}
\end{equation}
The estimation of the true parameters are denoted by $\hat{a}_{\mathcal{Z}}$ and $ \hat{b}_{\mathcal{Z}}$, solving Eq.~\ref{eq:par}, we can get 
\begin{equation}
\begin{aligned}\label{eq:parae}
& \hat{a}_{\mathcal{Z}} = \frac{\frac{1}{\mathcal{N_Z}} \sum_{z=1}^{\mathcal{N_Z}} L_V^z L_W^z - \overbar{L_V^\mathcal{Z}}\overbar{L_W^\mathcal{Z}}}{\frac{1}{\mathcal{N_Z}}\sum_{z=1}^{\mathcal{N_Z}} L_V^z  - \overbar{L_V^\mathcal{Z}}^2},  \quad \quad
\hat{b}_{\mathcal{Z}} = \overbar{L_W^\mathcal{Z}} - \hat{a}_{\mathcal{Z}} \overbar{L_V^\mathcal{Z}}
\end{aligned}
\end{equation}
where $\mathcal{Z}$ can be either $\mathcal{S}$ or $\mathcal{T}$, $\overbar{L_V^\mathcal{Z}}$ and $\overbar{L_W^\mathcal{Z}}$ are the mean of $L_V^\mathcal{Z}$ and $L_W^\mathcal{Z}$, respectively. Therefore, we are able to model the fitting lines of two domains via $ L_W^\mathcal{S} = \hat{a}_\mathcal{S} L_V^\mathcal{S} +\hat{b}_\mathcal{S}$ and $ L_W^\mathcal{T} = \hat{a}_\mathcal{T} L_V^\mathcal{T} +\hat{b}_\mathcal{T}$. In the following section, we present the distribution alignment with the estimated key coefficients $(\hat{a}_\mathcal{S},\hat{b}_\mathcal{S}, \hat{a}_\mathcal{T},\hat{b}_\mathcal{T})$, which are the slope and intercept of the source and the target domain, respectively.

\subsubsection{Marginal adaptation loss}
Before the alignment process, the marginal distribution of the two domains may partially overlap. To learn a separable geometric structure of marginal or global distribution, we consider the estimated parameters of two domains. It can be reached via maximizing the similarities between slope and intercept of two domains, which can be achieved by minimizing the following loss function:   
\begin{equation}\label{eq:mar}
  \mathcal{L_M} =  || \hat{a}_\mathcal{S} -\hat{a}_\mathcal{T}||_F^2 +  \gamma ||\hat{b}_\mathcal{S} - \hat{b}_\mathcal{T}||_F^2,
\end{equation}
where $\mathcal{M}$ denotes marginal distribution, $||\cdot||_F$ is the Frobenius norm and $\gamma$ balances the scale between two terms. The first term enforces small differences of slope between two domains, which is equivalent to minimizing the marginal/global angle ($\theta_\mathcal{M}$) between two fitting lines as in Eq.~\ref{eq:mtheta}.
\begin{equation}\label{eq:mtheta}
\theta_\mathcal{M} = \arccos{\frac{\hat{a}_\mathcal{S} \cdot \hat{a}_\mathcal{T}}{|\hat{a}_\mathcal{S}| \cdot |\hat{a}_\mathcal{T}|}}
\end{equation}
We can reformulate Eq.~\ref{eq:mar} as follows:
\begin{equation}\label{eq:mar2}
  \mathcal{L_M} =  \theta_\mathcal{M}  + \gamma  \mathcal{B}_\mathcal{M},
\end{equation}
where $\mathcal{B}_\mathcal{M} = ||\hat{b}_\mathcal{S} - \hat{b}_\mathcal{T}||_F^2$ represents the marginal intercept difference between two domains.
The marginal adaptation loss in Eq.~\ref{eq:mar2} first minimizing the angle of the two fitting lines, which is similar to rotating two fitting lines and leads to the same slope. It then minimizes the estimated intercepts of the two lines, which is equivalent to translating $\hat{b}_\mathcal{S}$ to $\hat{b}_\mathcal{T}$. As shown in Fig.~\ref{fig:model}, there is a 2-dimensional space with 2 classes. Let $\{1, 2\}$ be the labels of ``triangle" and ``cross".  According to the goal of Eq.~\ref{eq:mar2}, the marginal distribution alignment of the two domains can be achieved by finding a minimal $\theta_\mathcal{M}$ and intercept difference $\mathcal{B}_\mathcal{M}$.  

\subsubsection{Conditional adaptation loss}
Eq.~\ref{eq:mar2} can only minimize the marginal distribution divergences between the two domains. The conditional distributions are not aligned. Therefore, we need to design a conditional adaptation loss to minimize the conditional distribution divergences between the two domains. Since there are no labels in the target domain, the conditional distribution alignment is facilitated by soft pseudo-labels for the target domain. Given the trained classifier $\mathcal{F}$ in Eq.~\ref{eq:lc}, we can get the dominant predicted class for each target sample as $ \mathcal{Y}_\mathcal{T_P}^j = \text{argmax}\ \mathcal{F}(G(\mathcal{X}_{\mathcal{T}}^{j}))$. We hence get the label information for the latent target samples $G(\mathcal{X}_{\mathcal{T}})$ and these soft pseudo-labels can be refined via minimizing Eq.~\ref{eq:lc} and Eq.~\ref{eq:mar2}. To estimate the categorical slope and intercept, we modified Eq.~\ref{eq:parae} as:
\begin{equation}
\begin{aligned}
& \hat{a}_{\mathcal{Z}}^c = \frac{\frac{1}{\mathcal{N}_\mathcal{Z}^c} \sum_{z=1}^{\mathcal{N}_\mathcal{Z}^c} L_V^z L_W^z - \overbar{L_V^{\mathcal{Z}^c}}\overbar{L_W^{\mathcal{Z}^c}}}{\frac{1}{\mathcal{N}_\mathcal{Z}^c}\sum_{z=1}^{\mathcal{N}_\mathcal{Z}^c} L_V^z  - \overbar{L_V^{\mathcal{Z}^c}}^2} \quad \quad
\hat{b}_{\mathcal{Z}}^c = \overbar{L_W^{\mathcal{Z}^c}} - \hat{a}_{\mathcal{Z}}^c \overbar{L_V^{\mathcal{Z}^c}}
\end{aligned}
\end{equation}
where $\mathcal{Z}$ can be either $\mathcal{S}$ or $\mathcal{T}$, $\mathcal{N}_\mathcal{Z}^c$ is number of samples in each class $c$. $\overbar{L_V^{\mathcal{Z}^c}}$ and $\overbar{L_W^{\mathcal{Z}^c}}$ are the mean of $L_V^{\mathcal{Z}^c}$ and $L_W^{\mathcal{Z}^c}$ of class $c$. $L_V^z$ and $L_W^z$ naturally become one of the elements of $L_V^{\mathcal{Z}^c}$ and $L_W^{\mathcal{Z}^c}$. Therefore, we are able to model the categorical fitting lines of two domains with $ L_W^{\mathcal{S}^c} = \hat{a}_\mathcal{S}^c L_V^{\mathcal{S}^c} +\hat{b}_{\mathcal{S}}^c$ and $ L_W^{\mathcal{T}^c} = \hat{a}_\mathcal{T}^c L_V^{\mathcal{T}^c} +\hat{b}_\mathcal{T}^c$.  Similar to Eq.~\ref{eq:mar}, the conditional adaptation  loss is formulated as:
\begin{equation}\label{eq:con}
  \mathcal{L_C} =  \frac{1}{C}\sum_{c=1}^{C}|| \hat{a}_\mathcal{S}^c -\hat{a}_\mathcal{T}^c||_F^2 +  \gamma \frac{1}{C}\sum_{c=1}^{C}||\hat{b}_\mathcal{S}^c - \hat{b}_\mathcal{T}^c||_F^2. 
\end{equation}
Specifically, the estimated categorical slope and intercept are based on predicted pseudo-labels. Intuitively, the first term can also be regarded as minimizing the categorical angle $\theta_\mathcal{C}^c$ between fitting lines as follows.
\begin{equation}
\theta_\mathcal{C}^c = \arccos{\frac{\hat{a}_\mathcal{S}^c \cdot \hat{a}_\mathcal{T}^c}{|\hat{a}_\mathcal{S}^c| \cdot |\hat{a}_\mathcal{T}^c|}}
\end{equation}
Hence, we can rewrite Eq.~\ref{eq:con} as:
\begin{equation}\label{eq:con2}
  \mathcal{L_C} =  \frac{1}{C}\sum_{c=1}^{C} (\theta_\mathcal{C}^c  +   \gamma \mathcal{B}_\mathcal{C}^c), 
\end{equation}
where $\mathcal{B}_\mathcal{C}^c = ||\hat{b}_\mathcal{S}^c - \hat{b}_\mathcal{T}^c||_F^2$ denotes the conditional intercept difference between two domains of each class $c$. Therefore, the conditional adaptation loss considers each class, and minimizes categorical angle and intercept differences; it is naturally similar to performing categorical fitting line rotation and translation. As shown in Fig.~\ref{fig:model},  let $\theta_\mathcal{C}^1$ be the estimated angle between source and target fitting lines of ``triangle", and $\theta_\mathcal{C}^2$ be the estimated angle of ``cross".  According to Eq.~\ref{eq:con2}, the conditional distribution alignment of the two domains can be achieved by seeking minimal  $\theta_\mathcal{C}^1, \theta_\mathcal{C}^2$ and intercept difference ($\mathcal{B}_\mathcal{C}^1=||\hat{b}_\mathcal{S}^1 - \hat{b}_\mathcal{T}^1||, \mathcal{B}_\mathcal{C}^2=||\hat{b}_\mathcal{S}^2 - \hat{b}_\mathcal{T}^2||$).

\subsection{Overall objective function}
We integrate all components and obtain the following overall objective function of DLSA as:
\begin{equation}\label{eq:all}
  \mathcal{L} = \argmin \ (\mathcal{L_S}  + (1-\alpha) \mathcal{L_M} + \alpha \mathcal{L_C}),
\end{equation}
where $\mathcal{L_S}$ is the cross-entropy loss of the classifier in the labeled source domain.  $\mathcal{L_M}$ and $\mathcal{L_C}$ represent the marginal and conditional adaptation loss, respectively. $\alpha$ is the penalty parameter to balance conditional and marginal distribution. 

The overall training procedure is straightforward.  We first train the labeled source domain and unlabeled target domain using Eqs.~\ref{eq:lc} and~\ref{eq:mar2} to reduce the marginal distribution discrepancy between two domains. We then generate pseudo-labels ($\mathcal{Y_{T_P}}$) for the target domain with the trained classifier $\mathcal{F}$, and minimize the conditional adaptation loss using Eq.~\ref{eq:con2}. Finally, we repeat the previous two steps until the loss function in Eq.~\ref{eq:all} has converged.

\section{Experiments}
\subsection{Datasets}
\textbf{Office + Caltech-10} \cite{gong2012geodesic} has 2,533 images in four domains: Amazon (A), Webcam (W), DSLR (D), and Caltech (C) from ten classes. In experiments, A$\shortrightarrow$W represents transferring knowledge from domain A to domain W. We evaluate twelve tasks in this dataset. \textbf{Office-31} \cite{saenko2010adapting} has 4,110 images from three  domains: Amazon (A), Webcam (W), and DSLR (D) in 31 classes. We try six tasks in the Office-31 dataset. The \textbf{Office-Home}~\cite{venkateswara2017deep} dataset contains 15,588 images from four domains: Art (Ar), Clipart (Cl), Product (Pr) and Real-World (Rw) in 65 classes. We also test twelve tasks in this dataset. \textbf{VisDA-2017}~\cite{peng2017visda} is a particularly challenging dataset due to a large domain-shift between the synthetic images (152,397 images from VisDA) and the real images (55,388 images from COCO) in 12 classes. We test our model on the setting of synthetic-to-real as the source-to-target domain and report the accuracy of each category.  

\subsection{Implementation details}
We implement our approach using PyTorch with an Nvidia GeForce 1080 Ti GPU and extract features for the three datasets from a fine-tuned ResNet50 network~\cite{he2016deep}, which is a neural network well trained on the Imagenet dataset~\cite{krizhevsky2012imagenet}. The 1,000 features are then extracted from the last fully connected layer for the source and target features~\cite{zhang2019modified,zhang2020impact,zhang2020domain}. In the feature extractor $G$, the outputs of the first two Linear layers are 512, and the output of the last Linear layer is the number of classes in each dataset. The learning rate = 0.001, batch size = 32 and number of iterations = 300. 

During training, to balance the scale between slope and intercept, we return the value of angle $\theta_\mathcal{M}$ and $\theta_\mathcal{C}^c$ in radians, which is in the range of $[0,\pi]$.  The optimal parameters are $\alpha =0.2$, and $\gamma = 0.1$, and $\alpha \in \{0.1, 0.2, \cdots, 0.9\}$, while  $\gamma$ is selected from $\{0.1, 0.2, \cdots, 1\}$ based on the parameter analysis in supplemental material.\footnote{Source code is available at: \url{https://github.com/YoushanZhang/Transfer-Learning/tree/main/Code/Deep/DLSA}.} We also compare our results with 19 state-of-the-art methods (including both traditional methods and deep neural networks).

\begin{table*}[h]
\scriptsize
\begin{center}
\captionsetup{font=scriptsize}
\caption{Accuracy (\%) on Office + Caltech-10 (based on ResNet50)}
\vspace{+0.2cm}
\setlength{\tabcolsep}{+0.3mm}{
\begin{tabular}{rccccccccccccc}
\hline \label{tab:OC+10}
Task & C$\shortrightarrow$A &  C$\shortrightarrow$W & C$\shortrightarrow$D & A$\shortrightarrow$C & A$\shortrightarrow$W & A$\shortrightarrow$D & W$\shortrightarrow$C & W$\shortrightarrow$A & W$\shortrightarrow$D & D$\shortrightarrow$C & D$\shortrightarrow$A & D$\shortrightarrow$W & \textbf{Ave.}\\
\hline
\textbf{GSM}~\cite{zhang2019transductive} & 96.0 &	95.9&	96.2&	94.6 &	89.5&	92.4&	94.1&	95.8&	\textbf{100}	& 93.9&	95.1&	98.6&	95.2\\
\textbf{JGSA}~\cite{zhang2017joint} & 95.1&	97.6&	96.8&	93.9&	94.2&	96.2&	95.1&	95.9&	\textbf{100}	& 94.0 & 	\textbf{96.3} &	99.3&	96.2\\
\textbf{MEDA}~\cite{wang2018visual}	 & 96.3 &	98.3&	96.2&	94.6 &	99.0	&\textbf{100}	&94.8 &	\textbf{96.6}&	\textbf{100} & 93.6&	96.0	& 99.3&	97.0\\
\hline
\hline
DDC~\cite{tzeng2014deep}	&	91.9&	85.4&	88.8&		85.0&	86.1&	89.0&	78.0&	83.8&		\textbf{100}	&	79.0	&	87.1&	97.7	&	86.1\\
DCORAL~\cite{sun2016deep}	&	89.8&	97.3 &	91.0	&	91.9 &	\textbf{100}	& 90.5 &	83.7&	81.5&		90.1	& 88.6	&	80.1&	92.3&		89.7\\
DAN~\cite{long2015learning} 	&	92.0&	90.6&	89.3	&	84.1&	91.8&	91.7&	81.2&	92.1&		\textbf{100}	&	80.3	&	90.0 &	98.5&		90.1\\
RTN~\cite{long2016unsupervised}  &93.7 & 96.9 &94.2 &88.1 &95.2 & 95.5& 86.6& 92.5& \textbf{100} & 84.6& 93.8 & 99.2 &93.4 \\
MDDA~\cite{rahman2020minimum} &93.6 & 95.2 &93.4 &89.1 &95.7 & 96.6& 86.5&94.8 & \textbf{100} & 84.7& 94.7 & \textbf{99.4 }& 93.6\\
 \hline
  \hline
\textbf{DLSA}& \textbf{96.6	}& \textbf{98.6}	& \textbf{98.1}	& \textbf{95.4}	&98.9	&\textbf{100}	&\textbf{95.3} &	\textbf{96.6} &	\textbf{100}& \textbf{95.1} &	96.2	& 98.3  &	\textbf{97.4}\\
  \hline
\end{tabular}}
\vspace{-0.5cm}
\end{center}
\end{table*}

\begin{table*}[h!]
\begin{center}
\scriptsize
\captionsetup{font=scriptsize}
\caption{Accuracy (\%) on Office-Home dataset (based on ResNet50)}
\vspace{+0.2cm}
\setlength{\tabcolsep}{+0.5mm}{
\begin{tabular}{rccccccccccccc}
\hline \label{tab:OH}
Task & Ar$\shortrightarrow$Cl &  Ar$\shortrightarrow$Pr & Ar$\shortrightarrow$Rw & Cl$\shortrightarrow$Ar & Cl$\shortrightarrow$Pr & Cl$\shortrightarrow$Rw & Pr$\shortrightarrow$Ar & Pr$\shortrightarrow$Cl & Pr$\shortrightarrow$Rw & Rw$\shortrightarrow$Ar & Rw$\shortrightarrow$Cl & Rw$\shortrightarrow$Pr & \textbf{Ave.}\\
\hline
\textbf{GSM}~\cite{zhang2019transductive} &49.4&	75.5&	80.2&	62.9&	70.6&	70.3&	65.6&	50.0&	80.8&	72.4&	50.4&	81.6&	67.5 \\
\textbf{JGSA}~\cite{zhang2017joint} &45.8&	73.7&	74.5&	52.3&	70.2&	71.4&	58.8&	47.3&	74.2&	60.4&	48.4&	76.8&	62.8\\
\textbf{MEDA}~\cite{wang2018visual} & 49.1&	75.6&	79.1&	66.7&	77.2&	75.8&	68.2&	50.4&	79.9&	71.9&	53.2&	82.0&	69.1\\
\hline
\hline
DANN~\cite{ghifary2014domain} 	& 45.6	& 59.3& 	70.1& 	47.0& 	58.5& 	60.9& 	46.1& 	43.7& 	68.5& 	63.2& 	51.8& 	76.8& 	57.6\\
JAN~\cite{long2017deep}		& 45.9& 	61.2& 	68.9& 	50.4& 	59.7& 	61.0& 	45.8& 	43.4& 	70.3& 	63.9& 	52.4& 	76.8& 	58.3\\
CDAN-M~\cite{long2018conditional}	& 50.6& 	65.9& 	73.4& 	55.7& 	62.7& 	64.2& 	51.8& 	49.1& 	74.5& 	68.2& 	56.9& 	80.7& 	62.8\\
TAT~\cite{liu2019transferable} &51.6  &69.5 & 75.4 & 59.4 & 69.5 & 68.6 & 59.5 &50.5 &76.8 &70.9 &56.6 &81.6 &65.8 \\
ETD~\cite{li2020enhanced} &51.3 & 71.9& 85.7& 57.6 &69.2 &73.7 &57.8 &51.2 &79.3 &70.2 &57.5 &82.1 &67.3 \\
TADA~\cite{wang2019transferable} & 53.1 &72.3& 77.2& 59.1 &71.2& 72.1& 59.7& 53.1& 78.4 &72.4& \textbf{60.0} &82.9& 67.6 \\
SymNets~\cite{zhang2019domain} & 47.7 & 72.9 & 78.5 & 64.2  & 71.3  &74.2  & 64.2  & 48.8 &  79.5&  74.5 &52.6 & 82.7& 67.6 \\
DCAN~\cite{li2020domain} &54.5 &75.7 &81.2 &67.4 &74.0 &76.3 &67.4 &52.7 &80.6 &74.1 &59.1 &83.5 &70.5\\ 
RSDA~{\scriptsize \cite{gu2020spherical}} & 53.2 & 77.7 & 81.3 &  66.4 &  74.0 &  76.5 &  67.9 &  53.0 &  82.0 &  75.8 &  57.8 &  85.4 &  70.9 \\
SPL~{\scriptsize \cite{wang2020unsupervised}} &54.5 &77.8 & 81.9  & 65.1 & 78.0 & 81.1 & 66.0 & 53.1 & 82.8 & 69.9 & 55.3 & 86.0 & 71.0 \\
ESD~\cite{zhang2021enhanced} & 53.2 &	 75.9  &	82.0  &	\textbf{68.4} &	 \textbf{79.3}  & \textbf{79.4} &  69.2 &	 54.8 &	81.9 &	74.6 &	56.2  &	\textbf{83.8} & 71.6 \\
SHOT~{\scriptsize \cite{liang2020we}} &57.1 & 78.1 & 81.5 & 68.0 & 78.2 & 78.1 & 67.4 & 54.9 &82.2 &73.3 & 58.8 &  84.3 & 71.8 \\
\hline
\hline
\textbf{DLSA}& 56.3 &	79.4  &	82.5  &	67.4 &	78.4  & 78.6 & \textbf{69.4} &	54.5 &	82.1 &	\textbf{75.3} &	56.4  &	83.7 & 	71.7 \\
\textbf{SHOT+DLSA} &\textbf{57.6} &\textbf{80.1} & \textbf{82.7} & \textbf{68.4} & 78.9 &\textbf{79.4} & \textbf{69.4} & \textbf{55.1} & \textbf{82.4} &  \textbf{75.3} & 59.1 & \textbf{83.8} &\textbf{72.7} \\
\hline
\end{tabular}}
\vspace{-0.6cm}
\end{center}
\end{table*}

\begin{table} [h!]
	\scriptsize
	\begin{minipage}{0.5\linewidth}
		\scriptsize
\captionsetup{font=scriptsize}
		\caption{Accuracy (\%) on Office-31 (ResNet50)}
		\vspace{+0.2cm}
		\centering
		\resizebox{0.9\textwidth}{!}{%
\setlength{\tabcolsep}{+0.1mm}{		
\begin{tabular}{rcccccccc|c|c|c|c|c|c|c|c|}
\hline \label{tab:O31}
Task & A$\shortrightarrow$W &  A$\shortrightarrow$D & W$\shortrightarrow$A & W$\shortrightarrow$D & D$\shortrightarrow$A & D$\shortrightarrow$W  & \textbf{Ave.}\\
\hline
\textbf{GSM}~{\scriptsize \cite{zhang2019transductive}} &85.9&	84.1&	75.5&	97.2&	73.6&	95.6&	85.3 \\
\textbf{JGSA}~{\scriptsize \cite{zhang2017joint}} &89.1&	91.0&	77.9&	\textbf{100} &	77.6&	98.2&	89.0 \\
\textbf{MEDA}~{\scriptsize \cite{wang2018visual}} &91.7&	89.2&	77.2&	97.4&	76.5&	96.2&	88.0 \\
\hline
\hline
ADDA~{\scriptsize \cite{tzeng2017adversarial}}	&86.2	&77.8  &68.9	&98.4 &	69.5 &	96.2	& 82.9\\
JAN~{\scriptsize \cite{long2017deep}}&	85.4 &	84.7	&70.0 &	99.8	&68.6 &	97.4 &	84.3\\
DMRL~{\scriptsize \cite{wu2020dual}} &90.8 &93.4 & 71.2  &\textbf{100} &73.0  &99.0  & 87.9 \\
TAT~{\scriptsize \cite{liu2019transferable}} & 92.5 & 93.2 & 73.1 & \textbf{100}& 73.1 & \textbf{99.3} & 88.4 \\
TADA~{\scriptsize \cite{wang2019transferable}} &94.3 & 91.6  & 73.0  &99.8  & 72.9 & 98.7 & 88.4 \\
{\scriptsize SymNets~\cite{zhang2019domain}\par}& 90.8& 93.9  &72.5  & \textbf{100} & 74.6 & 98.8& 88.4 \\
SHOT~{\scriptsize \cite{liang2020we}} &90.1  & 94.0  & 74.3 & 99.9 & 74.7 & 98.4 & 88.6 \\
SPL~{\scriptsize \cite{wang2020unsupervised}} &  92.7 & 93.0  & 76.8 & 99.8 & 76.4 & 98.7  & 89.6 \\
CAN~{\scriptsize \cite{kang2019contrastive}} & 94.5 & 95.0  &77.0   &99.8  & 78.0 & 99.1 &  90.6 \\
RSDA~{\scriptsize \cite{gu2020spherical}} & 96.1  & 95.8  & 78.9 & 100.0  & 77.4  & 99.3 & 91.3 \\
\hline
\hline
\textbf{DLSA} & \textbf{95.2} &	\textbf{96.2} &	\textbf{80.4} &	99.2&	\textbf{80.7} &	98.0 &	\textbf{91.6}	\\
\hline
\end{tabular}}}
	\end{minipage}
	\hspace{+0.04cm}
	\scriptsize
\captionsetup{font=scriptsize}
	\begin{minipage}{0.5\linewidth}
		\caption{Ablation experiments on Office-31  ($\mathcal{M}$: marginal ada-\\ptation loss, $\mathcal{C}$: conditional adaptation loss).}
		\vspace{+0.2cm}
		\centering
		\resizebox{0.96\textwidth}{!}{%
\setlength{\tabcolsep}{+0.1mm}{		
\begin{tabular}{llllllll}
\hline \label{tab:ab}
Task & A$\shortrightarrow$W &  A$\shortrightarrow$D & W$\shortrightarrow$A & W$\shortrightarrow$D & D$\shortrightarrow$A & D$\shortrightarrow$W  & \textbf{Ave.}\\
\hline
DLSA$-\mathcal{C/M}$  & 89.0 &	87.4 &	75.2 &	97.2 &	76.5 &	95.4 &	87.9\\
DLSA$-\mathcal{C}$ & 92.6 &	89.9 &	78.5 &	98.2&	78.9 &	96.6&	89.1\\
DLSA$-\mathcal{M}$ & 95.1 &	95.1 &	79.1 &	99.0 &	79.8 &	97.2&	91.0	\\
\hline
\hline
\textbf{DLSA} & \textbf{95.2} &	\textbf{96.2} &	\textbf{80.4} &	\textbf{99.2} &	\textbf{80.7} &	\textbf{98.0} &	\textbf{91.6}	\\
\hline
\end{tabular}}}
	\end{minipage}
	 \vspace{-0.3cm}
\end{table}

\begin{table*}[h!]
\begin{center}
\scriptsize
\captionsetup{font=scriptsize}
\caption{Accuracy (\%) on VisDA-2017 dataset (based on ResNet101)}
\vspace{+0.2cm}
\setlength{\tabcolsep}{+1.1mm}{
\begin{tabular}{rccccccccccccc}
\hline \label{tab:VisDA}
Task &  plane & bcycl &  bus &  car &  horse&  knife &  mcycl & person &  plant & sktbrd & train & truck & \textbf{Ave.}\\
\hline
Source-only~\cite{he2016deep} &  55.1 & 53.3 & 61.9 & 59.1 &  80.6 & 17.9 &  79.7 & 31.2 &  81.0& 26.5& 73.5& 8.5 & 52.4 \\
 MCD~\cite{saito2018maximum}	&  87.0 & 60.9&  83.7&  64.0&  88.9&  79.6&  84.7&  76.9&  88.6&  40.3&  83.0&  25.8&  71.9\\
 DMP~\cite{luo2020unsupervised}  & 92.1  & 75.0  & 78.9 & 75.5 & 91.2 & 81.9 & 89.0 & 77.2 & 93.3 & 77.4 & 84.8 & 35.1 & 79.3 \\
 DADA~\cite{tang2020discriminative} &  92.9 & 74.2 &  82.5 &  65.0 &  90.9 &  93.8 &  87.2 &  74.2 &  89.9 &  71.5 &  86.5  & 48.7 &   79.8 \\
 STAR~\cite{lu2020stochastic} &  95.0 &  84.0 &  84.6 &  73.0 &  91.6 & 91.8&  85.9 & 78.4 &  94.4 &  84.7 & 87.0 & 42.2 &  82.7 \\
SHOT~{\scriptsize \cite{liang2020we}} &94.3 & 88.5 & 80.1 & 57.3 & 93.1 & 94.9 &80.7 & 80.3 & 91.5 &89.1 &86.3 &58.2 &82.9\\
DSGK~\cite{zhang2021deep}&  95.7 & 86.3 & 85.8 & 77.3 & 92.3 & 94.9 & 88.5 & 82.9 & 94.9 & 86.5 & 88.1 & 46.8 & 85.0 \\
 CAN~\cite{kang2019contrastive} &  97.9 &  87.2 &  82.5 &  74.3 &  97.8 &  96.2 &  90.8 &  80.7 &  96.6 & 96.3 & 87.5 &  59.9 &  87.2 \\
\hline
\hline
 \textbf{DLSA}&    96.9 &  \textbf{89.2}  &  85.4 &  77.9 &  98.3 &  \textbf{96.9} &  91.3 & 82.6 &  \textbf{96.9} &  \textbf{96.5} & 88.3 &  60.8 &  88.4 \\
 \textbf{CAN+DLSA} & \textbf{98.1} & \textbf{89.2} & \textbf{86.8} & \textbf{79.3} & \textbf{98.5} & \textbf{96.9} & \textbf{92.0} & \textbf{83.2} & \textbf{96.9} & \textbf{96.5} & 88.9 & \textbf{61.4} & \textbf{89.0}\\
\hline
\end{tabular}}
\vspace{-0.5cm}
\end{center}
\end{table*}

\subsection{Results}
Tables~\ref{tab:OC+10}-\ref{tab:O31} display the results of Office + Caltech-10, Office-Home and Office-31 datasets.  For a fair comparison, we bold three re-implemented baselines (GSM~\cite{zhang2019transductive}, JGSA~\cite{zhang2017joint}, and MEDA~\cite{wang2018visual}) using the same extracted features as our model. Our DLSA model still surpasses all state-of-the-art methods in general (most notably in the challenging  Office-31 dataset and the Office-Home dataset). 
 
In the Office + Caltech-10 dataset, compared with the best baseline MEDA, which is tested using our features, the accuracy of our method increases by 0.4\% on average. Although the improvement is not large, we still achieve the highest accuracy so far. For Office-31, the average accuracy of DLSA is 91.6\%. It is superior to all other methods. If we focus on the difficult tasks W$\shortrightarrow$A and D$\shortrightarrow$A, DLSA shows substantially better transferring ability than other methods. Our model  has a particularly obvious improvement in the challenging Office-Home dataset. The average accuracy is 71.7\%, which exceeds most SOTA methods. Our model also outperforms SOTA models in the challenging large-scale VisDA-2017 dataset in Tab.~\ref{tab:VisDA}.  
Our DLSA model can be an additional component for other models. We conducted experiments on two challenging datasets (Office-Home and VisDA-2017) and find that the combination of previous models with our proposed DLSA achieves the highest performance in Tab.~\ref{tab:OH} and Tab.~\ref{tab:VisDA} (SHOT+DLSA and CAN+DLSA). These two results demonstrate the effectiveness of DLSA in improving existing SOTA UDA models. Therefore, these experimental results show that our model outperforms all comparison methods, which reveal the DLSA is applicable to different datasets. 

\vspace{-0.4cm}
\paragraph{{\bf Ablation study.}}
To isolate the effects of marginal adaptation loss $\mathcal{L_M}$ and conditional adaptation loss $\mathcal{L_C}$ on classification accuracy, we perform an ablation study by evaluating different variants of DLSA in Tab.~\ref{tab:ab}.  ``DLSA$-\mathcal{C/M}$” is implemented without marginal and conditional adaptation losses. It is a simple model, which only reduces the source risk without minimizing the domain discrepancy using Eq.~\ref{eq:lc}. ``DLSA$-\mathcal{C}$” reports results without performing the additional conditional adaptation loss. ``DLSA$-\mathcal{M}$” trains the labeled source domain and performs the conditional adaptation. As the number of loss functions increases, the robustness of our model keeps improving. We can also conclude that $\mathcal{L_C}$ is more important than $\mathcal{L_M}$ in improving the performance. Therefore, the proposed marginal and conditional loss functions are helpful in minimizing the target domain risk.

\begin{figure*}[t]
\centering     
\captionsetup{font=small}
\subfigure[Estimated angle $\theta$]{\label{fig:maro31}\includegraphics[width=60mm]{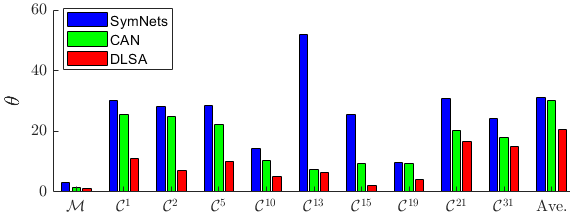}}
\subfigure[Estimated intercept differences $\mathcal{B}$]{\label{fig:co31}\includegraphics[width=60mm]{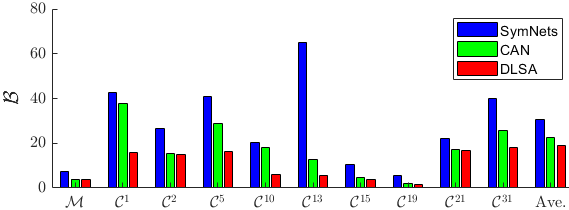}}
\caption{Least squares parameters comparison of task A$\shortrightarrow$D in the Office-31 dataset. (a) is the estimated angle $\theta$ of fitting lines between source and target domains. (b) is estimated intercept differences. DLSA consistently shows the smallest estimated parameters than  CAN~\cite{kang2019contrastive}, and  SymNets~\cite{zhang2019domain}. Ave. is the mean of 31 classes of $\theta_\mathcal{C}^c$ or $\mathcal{B}_\mathcal{C}^c$. }\label{fig:meth}
\vspace{-0.6cm}
\end{figure*}


\begin{figure*}[h!]
\centering
\captionsetup{font=small}
\includegraphics[width=1.\columnwidth]{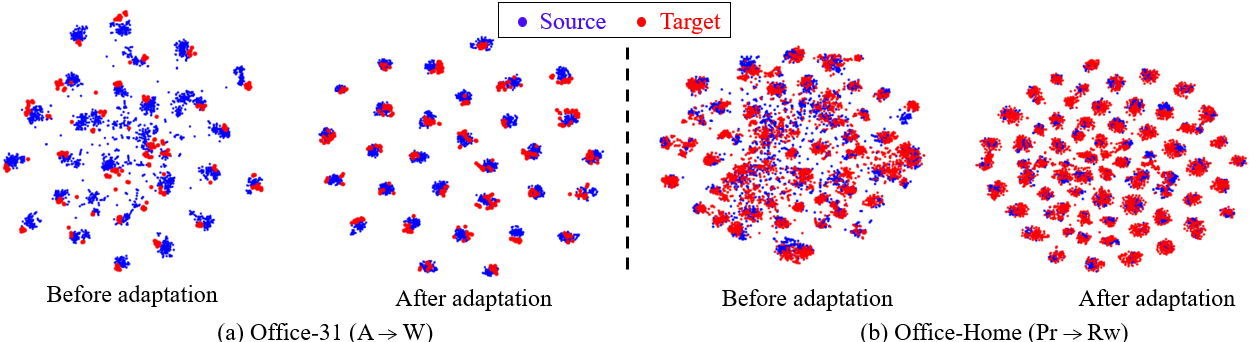}
\vspace{+0.02cm}
\caption{Visualization of learned features using a 2D t-SNE view of Office-31 and Office-Home dataset. }
\vspace{-0.3cm}
\label{fig:tsne}
\end{figure*}


To further investigate whether $\theta$ and $\mathcal{B}$ are minimized using DLSA, we also compare it with top baseline methods CAN~\cite{kang2019contrastive} and SymNets~\cite{zhang2019domain} with randomly selected task A$\shortrightarrow$D in Office-31 dataset in Fig.~\ref{fig:meth} (9 of 31 classes of conditional distribution are randomly reported). We can find that the estimated parameters $\theta$ and $\mathcal{B}$ are consistently smaller than the other two methods. Hence, we can find that DLSA can minimize marginal and conditional distribution discrepancy between two domains.        

\vspace{-0.4cm}
\paragraph{{\bf Feature Visualization.}}
To intuitively present adaptation performance, we utilize t-SNE to visualize the deep features of network activations in 2D space before and after distribution adaptation. Fig.~\ref{fig:tsne} shows two tasks:  (a) A$\shortrightarrow$W of Office-31 and (b) Pr$\shortrightarrow$Rw of Office-Home dataset. Apparently, the distributions of the two tasks become more discriminative after adaptation, while many categories are mixed up in the feature space before adaptation. It indicates that DLSA can learn more discriminative representations, which can significantly increase inter-class dispersion and intra-class compactness.

\vspace{-.1cm}
\section{Discussion}
\vspace{-.1cm}
In these experiments, DLSA always achieves the highest average accuracy. Therefore, the quality of our model exceeds that of SOTA methods, which reveals that DLSA can better learn domain invariant features and exceeds the frequently used MMD loss and CORAL loss. There are two reasons for this. First, DLSA can estimate the latent space distribution, which is parameterized by slope and intercept. Secondly, to reduce the discrepancy between domains in the latent feature space, we align the marginal distributions by reducing the angle and intercept differences between fitting lines of the domains. In addition, conditional distribution alignment is realized by generating soft pseudo labels. Then the categorical angle and intercept differences are minimized, which further supports agreement in label space. In terms of time, the major cost of estimating  slope and intercept is in Eq.~\ref{eq:parae}, which can be calculated in $\mathbb{O}(1)$. More comparisons can be found in supplementary material.

However, one disadvantage of our DLSA model is that we assume a linear relationship in the latent space. Other relationships may also improve the performance (e.g., polynomial/nonlinear).  Also, incorrect pseudo-labels may exist during training, affecting the quality of the fitting line as shown in Tab.~\ref{tab:para}. We can find that there are small differences of estimated parameters ($\theta_\mathcal{C}$ and $B_\mathcal{C}$) using pseudo-labels versus true labels from the target domain. Therefore, we can still improve the prediction using better pseudo-label generation methods. 
Removing outlier labels in the target domain might improve performance. We leave this to future work.

\begin{table*}[h]
\small
\captionsetup{font=small}
\begin{center}
\caption{Least squares estimated conditional parameters of task C$\shortrightarrow$W on Office + Caltech-10 dataset using pseudo and true target labels ($\mathcal{C}^c$: conditional distribution of each class $c$, where $c = \{1 , 2, 3, \cdots, 10\}$).}
\vspace{+0.25cm}
\setlength{\tabcolsep}{+0.5mm}{
\begin{tabular}{c||cccccccccc|ccc}
\hline \label{tab:para}
 Parameters &   $\mathcal{C}^1$ &  $\mathcal{C}^2$ &  $\mathcal{C}^3$  &  $\mathcal{C}^4$ &  $\mathcal{C}^5$  &  $\mathcal{C}^6$  &  $\mathcal{C}^7$  &  $\mathcal{C}^8$  &  $\mathcal{C}^9$  &  $\mathcal{C}^{10}$   &  \textbf{Ave.}\\
\hline
\hline        
 $\theta_\mathcal{C}^{\text{pseudo}}$ &	 29.786 &	 35.871  & 24.615 &	 16.451  &	 7.202 &	 11.924 &	 37.778 &	 0.355	&  33.137 &	 16.594 &	 21.371\\
 $\theta_\mathcal{C}^{\text{true}}$ &	 31.847 &	 36.301  & 26.824 &	 17.730 &	 7.074 &	 9.634 &	 36.169 &	 0.726	&  35.393 &	 18.824 &	 22.052\\
\hline  
 $\mathcal{B}_\mathcal{C}^{\text{pseudo}}$  &  33.137	&  9.875	&  17.656	&  3.622 &  3.626	& 0.250	& 15.820	& 3.097 &	 3.358 &	 16.237  & 	  10.668\\
 $\mathcal{B}_\mathcal{C}^{\text{true}}$  &  30.936	&  10.159	&  19.546	&  5.778 & 2.557	& 0.337	& 13.239	& 5.036 &	 4.088 &	 19.536  & 	  11.121\\
  \hline
\end{tabular}}
\vspace{-0.9cm}
\end{center}
\end{table*}

\vspace{-.1cm}
\section{Conclusion}
\vspace{-.1cm}
In this paper, we propose a novel unsupervised domain adaptation method, namely deep least squares adaptation. DLSA estimates the latent space distribution by finding the fitting lines of two domains. We then develop marginal and conditional adaptation loss to reduce domain discrepancy and learn domain invariant features. Experiential results illustrate the superiority of DLSA in modeling the source and target domain distributions, resulting in outstanding performances on benchmark datasets.

\bibliography{egbib}
\end{document}